\documentclass{article}
\usepackage{spconf,amsmath,graphicx,comment,multirow}

\title{Contextual Joint Factor Acoustic Embeddings}
\name{Yanpei Shi, Thomas Hain}
\address{Speech and Hearing Research Group\\
	Department of Computer Science, University of Sheffield, UK\\
\texttt{\{YShi30, t.hain\}@sheffield.ac.uk}}

\begin{document}
%
\maketitle
\begin{abstract}
Embedding acoustic information into fixed length representations is of interest for a whole range 
of applications in speech and audio technology. Two novel unsupervised approaches
to generate acoustic embeddings by modelling of acoustic context
are proposed. The first approach is a contextual joint factor
synthesis encoder, where the encoder in an encoder/decoder
framework is trained to extract joint factors from surrounding
audio frames to best generate the target output. The second
approach is a contextual joint factor analysis encoder, where
the encoder is trained to analyse joint factors from the source
signal that correlates best with the neighbouring audio. To
evaluate the effectiveness of our approaches compared to prior
work, two tasks are conducted- phone classification and speaker
recognition - and test on different TIMIT data sets. Experimental
results show that one of the proposed approaches outperforms
phone classification baselines, yielding a classification accuracy of
74.1\%. When using additional out-of-domain data for training,
an additional 3\% improvements can be obtained, for
both for phone classification and speaker recognition tasks.
\end{abstract}
\section{Introduction}
In recent years, word embeddings have been successfully used in natural language
processing (NLP), the most commonly known models are Word2Vec \cite{mikolov2013efficient}, Glove \cite{pennington2014glove} and BERT    \cite{devlin2018bert}. The reasons for such success are manifold. One key attribute of embedding methods is that word embedding models take into account context information, thereby allowing a more compact and manageable representation for words \cite{goldberg2014word2vec,li2015word}. Embeddings are widely applied in many downstream NLP tasks such as neural machine 
translation, dialogue system or text summarisation 
 \cite{cho2014learning,nallapati2016abstractive,dodge2015evaluating}, as well as in language 
modelling for speech recognition \cite{deena2017exploring}.

Embeddings of acoustic (and speech) signals are of more recent interest. The objective is to represent audio sequence information in compact form, replacing the raw audio data with one that contains only latent factors \cite{hsu2017unsupervised,bengio2014word}. The projection into such (latent) spaces should retain different attributes, such as phonemes, speaker properties, speaking styles, the acoustic background or the recording environment. Acoustic embeddings have been explored for a variety of speech tasks such as speech recognition \cite{settle2016discriminative}, speaker verification \cite{casanueva2016using} or voice conversion \cite{kamper2016deep}.
However, learning acoustic embeddings is challenging: attributes mentioned above, e.g. speaker properties and phonemes, operate at different levels of abstraction and are often strongly interdependent, and therefore are difficult to extract and represent in a meaningful form \cite{hsu2017unsupervised}.

For speech processing, \cite{chung2017learning,chung2018speech2vec,milde2018unspeech} also make use of context information to derive acoustic embeddings. \cite{chung2017learning,chung2018speech2vec} focus on learning word semantic 
representations from raw audio instead of signal properties such as phonemes and speaker properties. \cite{milde2018unspeech} focuses on learning speaker representations by 
modelling of context information with a Siamese network that discriminates whether a speech segment is the neighbourhood of a target segment or not. 

In this paper, two unsupervised approaches to generate acoustic embeddings using context modelling are proposed. Both methods make use of the variational auto-encoder framework as proposed in \cite{kingma2013auto} and both approaches aim to find joint latent variables for the target acoustic segments and its surrounding frames. In the first instance a representation is derived from surrounding audio frames that allows to predict current frame, thereby generating target audio from common factors. The encoder element of the 
associated auto-encoder is further referred to contextual joint factor synthesis (CJFS) encoder. In the second instance an audio frame is used to predict surrounding audio, which is further refereed to contextual joint factor analysis (CJFA) encoding. As shown in previous work variational auto-encoders can be used to derive latent variables such as speaker information and phonemes \cite{hsu2017unsupervised} robustly. In this work it is shown that including temporal information can further improve performance and robustness, for 
both phoneme classification and speaker identification tasks. Furthermore the use of additional unlabelled out-of-domain data can improve modelling for the proposed approaches. 
As outlined above, prior work has made use of surrounding audio in different forms. 
To the best of our knowledge, this work is the first to show that predicting surrounding audio allows for efficient extraction of latent factors in speech signals. 

The rest of paper is organised as follows: In \S\ref{sec:acem} related work is described, 
Methods for deriving acoustic embeddings, and context modelling methods in NLP, computer vision and speech are discussed. This is followed by the
description of the two approaches for modelling context as used in this work, in \S\ref{model architecture}. 
The experimental framework is described in \S\ref{sec:framework}, including the data organisation, baseline design and 
task definitions; in \S\ref{results} experiments results are shown and discussed. This is followed by the conclusions and future work in \S\ref{conclusion}. 

\vspace*{-2mm}
\section{Related Works} \label{sec:acem}
\vspace*{-2mm}
\subsection{Acoustic Embeddings}
Most interest in acoustic embeddings can be observed on acoustic word embeddings, i.e. projections 
that map word acoustics into a fixed size vector space. Objective functions are chosen to project 
different word realisations to close proximity in the embedding space. Different approaches were used in the literature - for both supervised and unsupervised learning. For the supervised case, \cite{bengio2014word} introduced a convolutional neural network (CNN) based acoustic word embedding system for speech recognition, where words that sound alike are close to each other in Euclidean distance. In their work, a CNN is used to predict a word from the corresponding acoustic signal, the output of the bottleneck layer is taken to be the embedding for the corresponding word. Further work used different network architectures to obtain acoustic word embeddings: 
\cite{settle2016discriminative} introduces a recurrent neural network (RNN) based approach. 

For the case that word boundary information is available but the word labels are unknown, \cite{kamper2016deep} proposed word similarity Siamese CNNs. These are used to minimise a distance function between representations of two instances of the same word type whilst at the same time
maximising the distance between two instances of different words.

Unsupervised approaches also exist. In \cite{hsu2017learning}, the authors chose phoneme and speaker classification tasks on TIMIT data to assess the quality of their embeddings - an approach replicated in the 
work presented in this paper. \cite{hsu2017unsupervised,hsu2018scalable} proposed an approach called factorised hierarchical variational auto-encoder, which introduces the concepts of global and local latent factors,
i.e. latent variables that are shared on the complete utterance, or latent variables that change within
the sequence, respectively. Results are again obtained using the same data and tasks as above. 

\subsection{Context Modelling} \label{sec:context}

Context information plays a fundamental role in speech processing. Phonemes could be influenced by surrounding frames through coarticulation - an effect caused by speed limitations and transitions in the movement of articulators \cite{ostry1996coarticulation}. Normally directly neighbouring phonemes have important impact on the sound realisation. Inversely, the surrounding phonemes also provide strong constraints on the  phoneme that can be chosen at any given point, subject to to lexical and language constraints. This effect is for example exploited in phoneme recognition, by use of phoneme $n$-gram models \cite{galescu2001bi}. Equivalently inter word dependency - derived from linguistic constraints - can be exploited, as is the case in computing word embeddings with the aforementioned  word2vec method \cite{mikolov2013efficient}. 
The situation differs for the global latent variables, such as speaker properties or acoustic 
environment information. Speaker properties remains constant - and environments can also be assumed 
stationary over longer periods of time. Hence these variables are common between among neighbouring
frames and windows. Modelling context information is helpful for identifying such
information \cite{higgins1994speaker}.

There are significant prior works that takes surrounding information into account to learn vector representations.
For text processing the Word2Vec \cite{mikolov2013efficient} model directly predicts the neighbouring words from target words or inversely. BERT model \cite{devlin2018bert} predicts the masked words in a sentence. This helps to capture the meanings of words \cite{goldberg2014word2vec}. In computer vision, \cite{pathak2016context} introduced an visual feature learning approach called context encoder, which is based on context based pixel prediction. Their model is trained to generate the contents of an image region from its surroundings.  In speech processing \cite{chung2017learning,chung2018speech2vec} proposed a sequence to sequence approach to predict surrounding segments of a target segment. However, the approach again aims at capturing word semantics from raw speech audio, words that has similar semantic meanings are nearby in Euclidean distance. \cite{milde2018unspeech} proposed an unsupervised acoustic embedding approach. In their approach, instead of directly estimating the neighbourhood frames of a target segment, a Siamese architecture is used to discriminate whether a speech segment is in the neighbourhood of a target segment or not. Furthermore, their approach only aims at embedding of speaker properties. To the best of our knowledge, work presented here is the first derive phoneme and speaker representations by temporal context prediction using acoustic data.

\section{Model Architecture}\label{model architecture}
\subsection{Variational Auto-Encoders} \label{sub:vae}
As shown in \cite{hsu2017learning}, variational auto-encoders (VAE) \cite{hsu2017unsupervised} can yield good representations in the latent space. One of the benefits is that the models allow to work with the latent distributions \cite{higgins2017beta,hsu2017unsupervised,kim2018disentangling}. In this work, VAE is used to model the joint latent factors between the target segments and its surroundings. 

Normal auto-encoders compressed the input data into latent code which is a point estimation of latent variables \cite{kingma2013auto}. Variational auto-encoder model defines a probabilistic generative process between the observation $x$ and the latent variable $z$. 
At the encoder step, the encoder provides an estimation of the latent variable $z$ given observation $x$ as $p(z|x)$.  The decoder finds the most likely reconstruction $\hat{x}$ subject to $p(\hat{x}|z)$. The latent variable estimation $p(z|x)$, or the probability density function thereof, has many interpretations, simply as encoding, or as latent state space governing the 
construction of the original signal. 

Computing $p(z|x)$ requires an estimate of the marginal likelihood $p(x)$ which is difficult to obtain in practice \cite{hsu2017unsupervised}. A recognition model $q(z|x)$ is used to approximate $p(z|x)$
KL divergence between $p(z|x)$ and $q(z|x)$, as shown in Eq \ref{vae_eq1}, is minimised \cite{kingma2013auto}. 
\begin{equation}\label{vae_eq1}
\begin{aligned}
&D_{KL}[q(z|x)||p(z|x)] = E[\log q(z|x) - \log \frac{p(x|z)p(z)}{p(x)}]\\
& = E[\log q(z|x) - \log p(x|z) - \log p(z)] +\log p(x)\\
& = E[\log p(x|z)] - D_{KL}[q(z|x)||p(z)] + p(x)
\end{aligned}
\end{equation}

From Eq \ref{vae_eq1}, the objective function for VAE training is derived in Eq \ref{px} \cite{kingma2013auto,hsu2017learning}:
\begin{equation}\label{px}
E_{q(z|x)}\log p(x|z) - D_{KL}(q(z|x)||p(z))
\end{equation}
where $E_{q(z|x)}log[p(x|z)]$ is also called the reconstruction likelihood and $ D_{KL}(q(z|x)||p(z))$ ensures the learned distribution $q(z|x)$ is close to prior distribution $p(z)$. 

\vspace*{-1mm}
\subsection{Proposed Model Architecture}\label{Context Prediction Method}
\vspace*{-1mm}
An audio signal is represented sequence of feature vectors $S=\{S_1,S_2,...S_T\}$, 
where $T$ is the length of the utterance. In the proposed method the concept of a 
target window is used, to which the embedding is related. A target window $X_t$ is a 
segment of speech representing features from $S_t$ to $S_{t+C-1}$, where $t \in \{1,2,..., T-C+1\}$ 
and $C$ denotes the target window size. The left neighbour window of the target window is defined as the segment between $S_{t-N}$ and $S_{t-1}$, and the segment between $S_{t+C}$ and $S_{t+C+N-1}$ represents the right neighbour window of the target window, with $N$ being the single sided neighbour window size. When $t-N <0$, the left neighbor window will be padded with zeros, and when $t+C+N-1>T$, the right neighbor window will be padded with zeros. 
The concatenation of left and right neighbour segments is further referred to $Y_t$. The proposed approach aims to find joint latent factors between target window segment $X_t$ and the concatenation of left and right neighbour window segments $Y_t$, for all segments. 
For convenience the subscript $t$ is dropped in following derivations where appropriate. Two different context use configurations can be used. 

\begin{figure}[h]
	\centering
	\includegraphics[height=10cm,width=9cm]{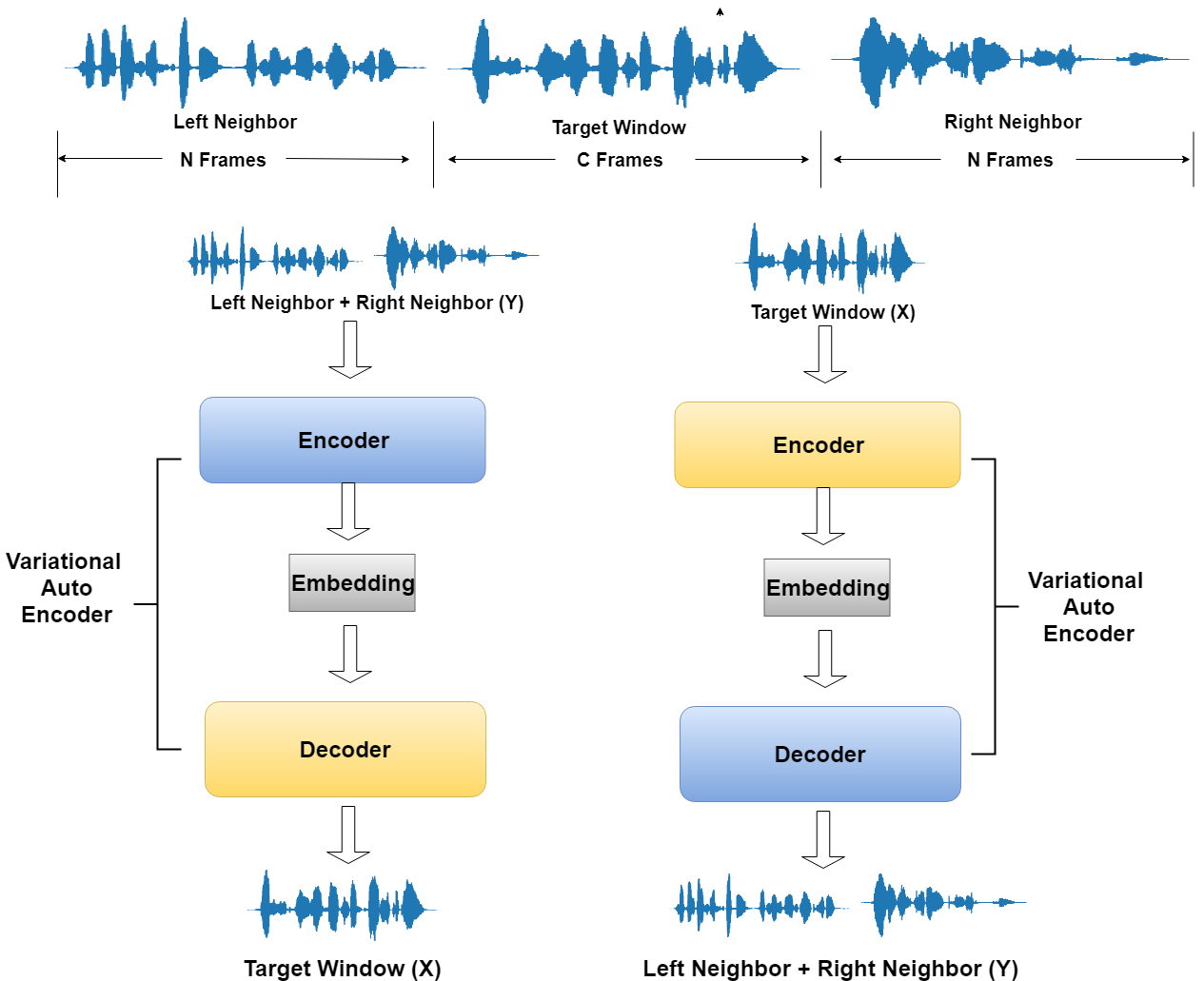}
	\vspace{-1mm}
	\caption{The architecture of CJFS (left) and CJFA (right). Both were built based on variational auto encoder. Embeddings were extracted on the bottleneck layer.} \label{context prediction}
\end{figure}
Figure \ref{context prediction} illustrate these two approaches. The audio signal is split into a sequence of left neighbour segment, target segment and right neighbour segment. In the first approach (left side on figure \ref{context prediction}), the concatenation of the left neighbour segment and right neighbour segment ($Y$) is input to a VAE model \cite{kingma2013auto}, and target window ($X$) is predicted. In the second approach (right side on figure \ref{context prediction}) the target window ($X$) is the input to a VAE model, and neighbour window ($Y$) is predicted. 

The first approach is referred to the contextual joint factor synthesis (CJFS) encoder as it 
aims to synthesise the target window $X$. Only factors common between input and output 
can form the basis for such prediction, and the encoded embedding can be considered a 
representation of these joint factors. Similar to the standard VAE formulations, 
the objective function of CJFS is given in Eq. \ref{loss_context}:
\begin{equation}\label{loss_context}
E_{q(z|Y)}log[p(X|z)] - KL(q(z|Y)||p(z))
\end{equation}
The first term represents the reconstruction likelihood between predicted target window segments and the neighbour window segments, and the second term denotes how similar
the learned distribution $q(z|Y)$ is to the prior distribution of $z$, $p(z)$. 

In practice, the reconstruction term is based on the mean squared error (MSE) 
between the true target segment and the predicted target segment. For the second term in Eq. \ref{loss_context}, samples for $p(z)$ are obtained from Gaussian distribution with zero mean and a variance of one ($p(z) \sim \mathcal N (0,1)$). 


The second approach is the contextual joint factor analysis (CJFA) encoder. The objective is to 
predict the temporal context $Y$ based on input from a single central segment $X$. 
Again joint factors between the three windows are obtained, and encoded in an 
embedding. The training
objective function of CJFA is represented by change of variables, as given in Eq \ref{loss_inver}.
\begin{equation}
\label{loss_inver}
E_{q(z|X)}log[p(Y|z)] - KL(q(z|X)||p(z))
\end{equation}


\section{Experimental Framework}\label{sec:framework}
\vspace*{-3mm}
\subsection{Data and Use} \label{data and use}
The TIMIT corpus is used for this work \cite{zue1990speech}.
TIMIT contains studio recordings from a large number of speakers with detailed phoneme segment information. Work in this paper makes use of the official training and test sets, 
covering in total 630 speakers with 8 utterances each. There is no speaker overlap 
between training and test set, which comprise of 462 and 168 speakers, respectively. 
All work presented here use of 80 dimensional Mel-scale filter bank coefficients. 
\vspace*{-3mm}
\subsection{Baseline}
\vspace*{-1mm}

The work on VAE in \cite{hsu2017learning} conducted experiments using
the TIMIT data set to learn acoustic embeddings. In particular the tasks of phone classification and speaker recognition 
were chosen. As work here is an extension of such work, the work in \cite{hsu2017learning} is used as the baseline, the experimentation is followed however with significant extensions (see Section \ref{evaluation}).

With guidance from 
the authors of the original work \cite{hsu2017learning}, our own implementation of VAE  
was created and compared with the published performance - yielding near identical results. 
The baseline performance for VAE based phone classification experiments in \cite{hsu2017learning} report an accuracy of 72.2\%. The re-implementation forming the basis for our work gave an accuracy of 72.0\%, a result that was considered to provide a credible basis for further work. 
This implementation then was also used as the basis for CJFS and CJFA,
as introduced in \S\ \ref{Context Prediction Method}. 



\begin{figure}[t]
	\centering
	\includegraphics[height=3cm,width=8.5cm]{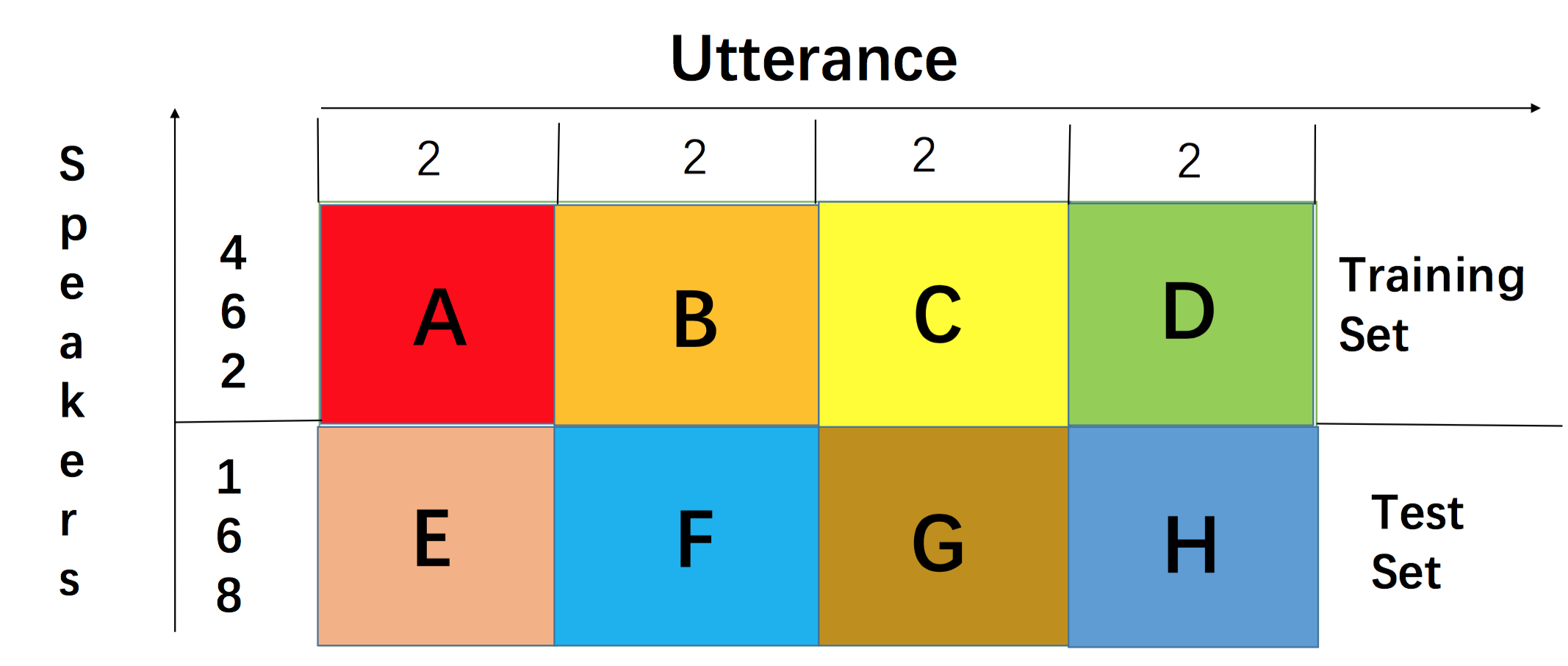}
	\vspace{-2mm}
	\caption{Data split of the TIMIT corpus for definition of data sets for speaker recognition. Training
		and test sets are split into 4 parts of 2 utterances each. Different combination of sets for training and test are used of different tasks.  
	} \label{fig:data}
\end{figure}

\vspace*{-3mm}
\subsection{Evaluation}\label{evaluation}
\vspace{-1mm}

For the assessment of embedded vector quality, this work also follows the same task types in \cite{hsu2017learning}, namely  phone classification and speaker recognition, with identical
task implementations as in the reference paper.

The phone classification implementation operates on segment level, using a 
convolutional network to obtain frame by frame posteriors which are then 
accumulated for segment decision (assuming frame independence). The phone class
with the highest segment posterior is chosen as output. It is important to note that phone classification differs from the widely reported phone recognition experiments on TIMIT. Classification uses phone boundaries which are assumed to be known. However, no context information is available, which is typically used in the recognition setups, by means of triphone models, or bigram language models.
Therefore the task is often more difficult than recognition.

An identical approach 
is used for speaker recognition. In this setting 3 different data sets are required:
a training set for learning the encoder models, a training set for learning 
the classification model, and an evaluation test set. 
For the phone classification task, both embedding and classification models are trained on the official TIMIT training set, and makes use of the provided phone boundary information. A fixed size window with a frame step size of one frame is used for all model training. As noted, phone classification makes no use of phone context, and no language model is applied.  
\begin{figure*}[h]
	\centering
	\includegraphics[height=5.3cm,width=17.5cm]{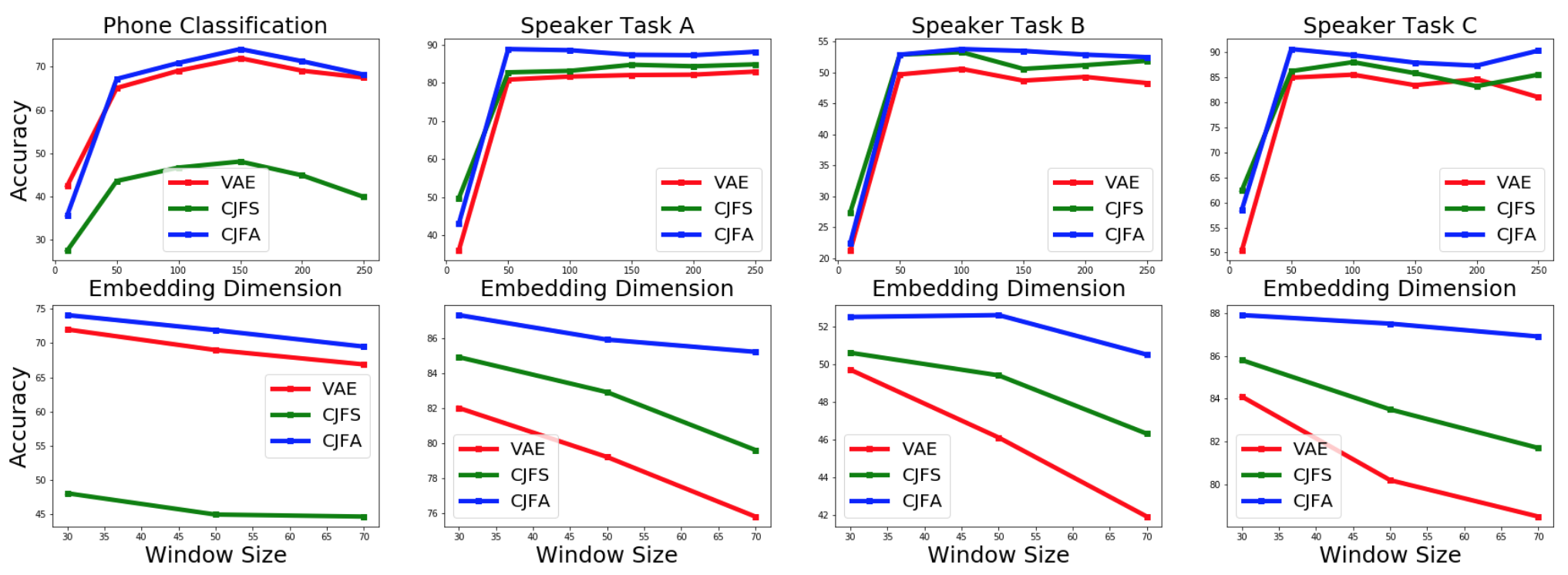}
	\caption{Phone classification accuracy, and speaker recognition accuracy for Tasks a,b,and c (as defined at \ref{evaluation}), when varying the embedding dimension (top row), 
		and window sizes (bottom row).}\label{compares}
\end{figure*}

\begin{figure*}[t]
	\centering
	\includegraphics[height=4.1cm,width=17cm]{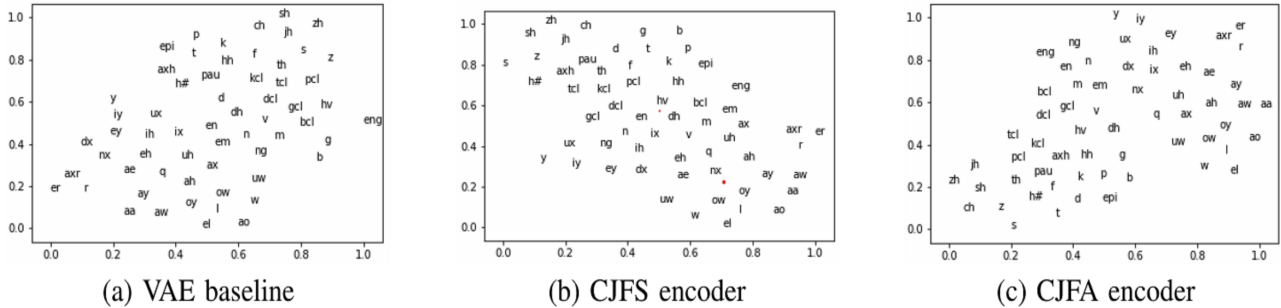}
	\vspace{-2mm}
	\caption{The t-SNE visualisation of phones in the test set for three models: (a): VAE baseline (b): CJFS encoder and (c): CJFA encoder 
	} \label{figure butterfly}
	\vspace*{-3mm}
\end{figure*}

For the purpose of speaker recognition, it is important to take into account the speaker overlap 
between training and testing. Thus three different task configurations are considered,
different to the setting in \cite{hsu2017learning}. 
As speakers between any of the datasets (training embeddings, training classifier and test) will cause a bias. Three different configurations (Tasks a,b,c) 
are used to assess this bias. Task a reflects the situation where both classifier and embedding are 
trained on the same data. As the task is to detect a speaker the speakers present in the test set need to be present in training. Task b represents a situation where classifier and embedding are trained on independent data sets, but with speaker overlap. Finally Task c represents complete independence in training data sets and no speaker overlap.  Table \ref{joint-overlap} summarises the relationships. 

\renewcommand\arraystretch{1.0}
\begin{table}[t]
	\centering
	\setlength{\tabcolsep}{2mm}{
		\begin{tabular}{ c c c c}
			\hline
			&Task a&Task b&Task c\\
			\hline
			Joint Training Sets &Yes& No &No\\
			Speaker Overlap&Yes&Yes&No\\
			\hline
		\end{tabular}
		\vspace{-1mm}
		\caption{Definition of training configurations a, b, and c.}
		\label{joint-overlap}}
	\vspace*{-1mm}
\end{table}

In order to achieve these configuration the TIMIT data was split. Fig. \ref{fig:data} illustrates the split of the data into 8 subsets (A--H). The TIMIT dataset contains speech from 462 speakers in  training and 168 speakers in the test set, with 8 utterances for each speaker. 
The TIMIT training and test set are split into 8 blocks, where each block contains 2 utterances
per speaker, randomly chosen. Thus each block A,B,C,D contains data from 462 speakers with 924 utterances taken from the training sets, and each block E,F,G,H contains speech from 168 test set speakers with 336 utterances.

For Task a training of embeddings and the classifier is identical, namely consisting of 
data from blocks (from A to G). The test data is the remainder, namely blocks (D+H). 
For Task b the training of embeddings and classifiers uses (A+B+E+F) and (C+G) respectively, while again using (D+H) for test. Task c keeps both separate: embeddings are trained on (A+B+C+D), classifiers on (E+G) and tests are conducted on (F+H). Note that H is part of all tasks, and that Task c is considerably easier as the number of speakers to separate is only 168, although training conditions are more difficult. 

\vspace{-2mm}
\subsection{Implementation} \label{sec:expt} 
\vspace{-2mm}
For comparison the implementation, follows the convolutional model structure as deployed in \cite{hsu2017learning}. Both VAE encoder and decoder contain three convolutional layers and one fully-connected layer with 512 nodes. In the first layer of encoder, 1-by-80 filters are applied, and 3-by-1 filters are applied on the following two convolutional layer (strides was set to 1 in the first layer and 2 in the rest two layers). The decoder has the symmetric architecture to the encoder.  
Each layer is followed by a batch normalisation layer \cite{ioffe2015batch} except for the embedding layer, which is linear. Leaky ReLU activation \cite{xu2015empirical} is used for each layer except for the embedding layer. The Adam optimizer \cite{kingmaadam} is used in training, with $\beta_1$ set to 0.95, $\beta_2$ to 0.999, and $\epsilon$ is $10^{-8}$. The initial learning rate is $10^{-3}$.
\vspace*{-3mm}
\section{Results and Discussion}\label{results}
\vspace{-3mm}
Table \ref{four models} shows phone classification and speaker recognition results for the three 
model configurations: the VAE baseline, the CJFS encoder and the CJFA encoder. In our experiments the window size was set to 30 frames, namely 10 frames for the target and 10 frames for left and right neighbours, and an embedding dimension of 150. This was used for both CJFS and CJFA models alike. 
Results show that the CJFA encoder obtains significantly better phone classification accuracy than the VAE baseline and also than the CJFS encoder. These results are replicated for speaker recognition tasks. The CJFA encoder performs better on all tasks than the VAE baseline by a significant margin. 
It is noteworthy that performance on Task b is generally significantly lower than for Task a, 
for reasons of training overlap but also smaller training set sizes. 

\renewcommand\arraystretch{1.0}
\begin{table}[h]
	\centering
	\setlength{\tabcolsep}{2mm}{
		\begin{tabular}{ c c c c c }
			\hline
			Model&Phone&Task a&Task b&Task c\\
			\hline
			
			VAE&72.0\%&82.0\%&49.7\%&84.1\%\\
			CJFS&48.1\%&84.9\%&50.2\%&85.8\%\\
			CJFA&74.1\%&87.3\%&52.2\%&87.9\%\\
			
			\hline
		\end{tabular}
		\caption{\% Phone classification and speaker recognition accuracy with three different model types. Embedding dimension is 150 and target window size is 10 frames, neighbour window sizes are 10 frames each.}
		\label{four models}}
\end{table}

To further explore properties of the embedding systems a change of window size ($N$) and embedding dimension ($K$) is explored. One might argue that modelling context effectively widens the input 
data access. Hence these experiments should explore if there is benefit in the structure beyond data size.
Graphs in Fig. \ref{compares} illustrate phone classification accuracy and speaker recognition performance for all three models under variation of latent size and window sizes. It is important 
to note that the target window size remains the same (10 frames) with an increase of $N$. Therefore
e.g. $N=70$ describes the target window size is 10 frames, and the other two neighbour windows have 30 frames at either side (30,10,30 left to right). Better speaker recognition results are consistently obtained with the CJFA encoder for any configuration with competitive performance, compared with the VAE baseline and also CJFS settings - and CJFS settings mostly outperform the baseline. However the situation for phone classification is different. It is not surprising to see CJFS perform poorly 
on phone classification as the target frame in not present in the input, therefore the embedding 
just does not have the phone segment information. However, as per speaker recognition results, speaker
information is retained. 

A variation of the window sizes to larger windows seems detrimental in almost all cases, aside from the more difficult Task b. This may be in part the effect of the amount of training data available, however it confirms that contextual models outperform the baseline VAE model configuration, generally, and in particular also with the same amount of input data for speaker recognition. It is also noticeable that the decline or variation as a function of window size is less pronounced for the CJFA case, implying increased stability. For phone classification the trade-off benefit for window size is less clear. 

For phone classification, increasing the embedding $K$ is helpful, but performance remains stable at $K=150$. Hence in all of the rest of our experiments, the embedding dimension is set to 150 for all of the rest configurations. For speaker recognition the observed variations are small. 

\renewcommand\arraystretch{1.0}
\begin{table}[h]
	\footnotesize
	\centering
	\setlength{\tabcolsep}{2mm}{
		\begin{tabular}{ c c c c c c}
			\hline
			&Data&Phone&Task a&Task b&Task c\\
			\hline
			
			VAE&TIMIT&72.0\%&82.0\%&49.7\%&84.1\%\\
			VAE+Lib&TIMIT+Lib&74.4\%  &87.6\%&57.3\%&87.4\%\\
			
			CJFS&TIMIT&48.1\%&84.9\%&50.2\%&85.8\%\\
			CJFS+Lib&TIMIT+Lib&52.4\%& 90.7\%&59.7\%&91.4\%\\
			
			CJFA&TIMIT&74.1\%& 87.3\%&52.2\%&87.9\%\\
			CJFA+Lib&TIMIT+Lib&\bfseries{76.3}\%&\bfseries{91.2}\%&\bfseries{62.4}\%&\bfseries{92.3}\%\\
			
			\hline
		\end{tabular}
		
		\caption{\% Phone classification and speaker recognition accuracies on TIMIT and LibriSpeech datasets (Lib represents LibriSpeech corpus.)}
		\label{more data}}
\end{table}

A further set of experiments investigated the use of out of domain data for improving 
classification in a completely unsupervised setting. The LibriSpeech corpus \cite{panayotov2015librispeech} 
was used in this case to augment the TIMIT data for training the embeddings only.
All other configurations an training settings are unchanged. Table \ref{more data} shows improvement after using additional out-of-domain data for training, except for in the case of CJFS and for phone classification. The improvement on all tasks with the simple addition of unlabelled audio data is 
remarkable. This is also true for the baseline, but the benefit of the proposed methods seems unaffected. The CJFA encoder performs better in comparison of the other two approaches and 
an absolute accuracy improvement of 7.9\% for speaker recognition Task b is observed. The 
classification tasks benefits from the additional data even though the labelled data remains the same.

To further evaluate the embeddings produced by the 3 models, visualisation using the t-SNE algorithm \cite{maaten2008visualizing} is a common approach, although interpretation is sometimes
difficult. Fig. \ref{figure butterfly} visualises the embeddings of phonemes in two-dimensional space, each phoneme symbol represents the mean vector of all of the embeddings belonging to the same phone class \cite{wu2018improving}. One can observe that the CJFA encoder appears to generate more meaningful embeddings than the other two approaches - as phonemes belonging to the same sound classes \cite{list2010sca} are grouped together in closer regions. The VAE baseline also has this behaviour but for example plosives are split and nasal separation seems less clear. Instead CJFS shows more confusion - as expected and explained above. 

\vspace{-3mm}
\section{Conclusion and Future Work}\label{conclusion}
\vspace{-3mm}
In this paper, two unsupervised acoustic embedding approaches to model the joint latent factors between the target window and neighbouring audio segments were proposed. Models are based on variational auto-encoders, which also constitute the baseline. In order to compare against the baseline models
are assessed using phone classification and speaker recognition tasks, on TIMIT, and with additional LibriSpeech data. Results show CJFA (contextual joint factor analysis) encoder performs significantly better in both phone classification and speaker recognition tasks compared with other two approaches. The 
CJFS (contextual joint factor synthesis) encoder performs close to CJFA in speaker recognition task, but poorer for phone classification. Overall a gain of up to 3\% relative on phone classification accuracy is observed, relative improvements on speaker recognition show 3--6\% gain. The proposed unsupervised approaches obtain embeddings and can be improved with unlabelled out-of-domain data, the classification tasks benefits even though the labelled data remains the same. Further work needs to 
expand experiments on larger data sets, phone recognition and more complex neural network architectures. 

\newpage
\bibliographystyle{IEEEbib}
\bibliography{strings,refs}

\begin{thebibliography}{10}

\bibitem{mikolov2013efficient}
Tomas Mikolov, Kai Chen, Greg~S Corrado, and Jeffrey Dean,
\newblock ``Efficient estimation of word representations in vector space,''
\newblock 2013.

\bibitem{pennington2014glove}
Jeffrey Pennington, Richard Socher, and Christopher Manning,
\newblock ``Glove: Global vectors for word representation,''
\newblock in {\em Proceedings of the 2014 conference on empirical methods in
  natural language processing (EMNLP)}, 2014, pp. 1532--1543.

\bibitem{devlin2018bert}
Jacob Devlin, Ming-Wei Chang, Kenton Lee, and Kristina Toutanova,
\newblock ``Bert: Pre-training of deep bidirectional transformers for language
  understanding,''
\newblock {\em arXiv preprint arXiv:1810.04805}, 2018.

\bibitem{goldberg2014word2vec}
Yoav Goldberg and Omer Levy,
\newblock ``word2vec explained: deriving mikolov et al.'s negative-sampling
  word-embedding method,''
\newblock {\em arXiv preprint arXiv:1402.3722}, 2014.

\bibitem{li2015word}
Yitan Li, Linli Xu, Fei Tian, Liang Jiang, Xiaowei Zhong, and Enhong Chen,
\newblock ``Word embedding revisited: A new representation learning and
  explicit matrix factorization perspective,''
\newblock in {\em Twenty-Fourth International Joint Conference on Artificial
  Intelligence}, 2015.

\bibitem{cho2014learning}
Kyunghyun Cho, Bart van Merrienboer, Caglar Gulcehre, Dzmitry Bahdanau, Fethi
  Bougares, Holger Schwenk, and Yoshua Bengio,
\newblock ``Learning phrase representations using rnn encoder--decoder for
  statistical machine translation,''
\newblock in {\em Proceedings of the 2014 Conference on Empirical Methods in
  Natural Language Processing (EMNLP)}, 2014, pp. 1724--1734.

\bibitem{nallapati2016abstractive}
Ramesh Nallapati, Bowen Zhou, Cicero dos Santos, Caglar Gulcehre, and Bing
  Xiang,
\newblock ``Abstractive text summarization using sequence-to-sequence rnns and
  beyond,''
\newblock in {\em Proceedings of The 20th SIGNLL Conference on Computational
  Natural Language Learning}, 2016, pp. 280--290.

\bibitem{dodge2015evaluating}
Jesse Dodge, Andreea Gane, Xiang Zhang, Antoine Bordes, Sumit Chopra, Alexander
  Miller, Arthur Szlam, and Jason Weston,
\newblock ``Evaluating prerequisite qualities for learning end-to-end dialog
  systems,''
\newblock {\em arXiv}, 2015.

\bibitem{deena2017exploring}
Salil Deena, Raymond~WM Ng, Pranava Madhyastha, Lucia Specia, and Thomas Hain,
\newblock ``Exploring the use of acoustic embeddings in neural machine
  translation,''
\newblock in {\em 2017 IEEE Automatic Speech Recognition and Understanding
  Workshop (ASRU)}. IEEE, 2017, pp. 450--457.

\bibitem{hsu2017unsupervised}
Wei-Ning Hsu, Yu~Zhang, and James Glass,
\newblock ``Unsupervised learning of disentangled and interpretable
  representations from sequential data,''
\newblock in {\em Advances in neural information processing systems}, 2017, pp.
  1878--1889.

\bibitem{bengio2014word}
Samy Bengio and Georg Heigold,
\newblock ``Word embeddings for speech recognition,''
\newblock in {\em Fifteenth Annual Conference of the International Speech
  Communication Association}, 2014.

\bibitem{settle2016discriminative}
Shane Settle and Karen Livescu,
\newblock ``Discriminative acoustic word embeddings: Tecurrent neural
  network-based approaches,''
\newblock in {\em 2016 IEEE Spoken Language Technology Workshop (SLT)}. IEEE,
  2016, pp. 503--510.

\bibitem{casanueva2016using}
Inigo Casanueva, Thomas Hain, Mauro Nicolao, and Phil Green,
\newblock ``Using phone features to improve dialogue state tracking
  generalisation to unseen states,''
\newblock in {\em Proceedings of the 17th Annual Meeting of the Special
  Interest Group on Discourse and Dialogue}, 2016, pp. 80--89.

\bibitem{kamper2016deep}
Herman Kamper, Weiran Wang, and Karen Livescu,
\newblock ``Deep convolutional acoustic word embeddings using word-pair side
  information,''
\newblock in {\em 2016 IEEE International Conference on Acoustics, Speech and
  Signal Processing (ICASSP)}. IEEE, 2016, pp. 4950--4954.

\bibitem{chung2017learning}
Yu-An Chung and James Glass,
\newblock ``Learning word embeddings from speech,''
\newblock {\em arXiv}, 2017.

\bibitem{chung2018speech2vec}
Yu-An Chung and James Glass,
\newblock ``Speech2vec: A sequence-to-sequence framework for learning word
  embeddings from speech,''
\newblock {\em Proc. Interspeech 2018}, pp. 811--815, 2018.

\bibitem{milde2018unspeech}
Benjamin Milde and Chris Biemann,
\newblock ``Unspeech: Unsupervised speech context embeddings,''
\newblock {\em Proc. Interspeech 2018}, pp. 2693--2697, 2018.

\bibitem{kingma2013auto}
Diederik~P Kingma and Max Welling,
\newblock ``Auto-encoding variational bayes,''
\newblock {\em arXiv preprint arXiv:1312.6114}, 2013.

\bibitem{hsu2017learning}
Wei-Ning Hsu, Yu~Zhang, and James Glass,
\newblock ``Learning latent representations for speech generation and
  transformation,''
\newblock {\em Proc. Interspeech 2017}, pp. 1273--1277, 2017.

\bibitem{hsu2018scalable}
Wei-Ning Hsu and James Glass,
\newblock ``Scalable factorized hierarchical variational autoencoder
  training,''
\newblock {\em Proc. Interspeech 2018}, pp. 1462--1466, 2018.

\bibitem{ostry1996coarticulation}
David~J Ostry, Paul~L Gribble, and Vincent~L Gracco,
\newblock ``Coarticulation of jaw movements in speech production: is context
  sensitivity in speech kinematics centrally planned?,''
\newblock {\em Journal of Neuroscience}, vol. 16, no. 4, pp. 1570--1579, 1996.

\bibitem{galescu2001bi}
Lucian Galescu and James~F Allen,
\newblock ``Bi-directional conversion between graphemes and phonemes using a
  joint n-gram model,''
\newblock in {\em 4th ISCA Tutorial and Research Workshop (ITRW) on Speech
  Synthesis}, 2001.

\bibitem{higgins1994speaker}
Alan~L Higgins,
\newblock ``Speaker verifier using nearest-neighbor distance measure,'' Aug.~16
  1994,
\newblock US Patent 5,339,385.

\bibitem{pathak2016context}
Deepak Pathak, Philipp Krahenbuhl, Jeff Donahue, Trevor Darrell, and Alexei~A
  Efros,
\newblock ``Context encoders: Feature learning by inpainting,''
\newblock in {\em Proceedings of the IEEE conference on computer vision and
  pattern recognition}, 2016, pp. 2536--2544.

\bibitem{higgins2017beta}
Irina Higgins, Loic Matthey, Arka Pal, Christopher Burgess, Xavier Glorot,
  Matthew Botvinick, Shakir Mohamed, and Alexander Lerchner,
\newblock ``beta-vae: Learning basic visual concepts with a constrained
  variational framework.,''
\newblock {\em Iclr}, vol. 2, no. 5, pp. 6, 2017.

\bibitem{kim2018disentangling}
Hyunjik Kim and Andriy Mnih,
\newblock ``Disentangling by factorising,''
\newblock in {\em International Conference on Machine Learning}, 2018, pp.
  2654--2663.

\bibitem{zue1990speech}
Victor Zue, Stephanie Seneff, and James Glass,
\newblock ``Speech database development at mit: Timit and beyond,''
\newblock {\em Speech communication}, vol. 9, no. 4, pp. 351--356, 1990.

\bibitem{ioffe2015batch}
Sergey Ioffe and Christian Szegedy,
\newblock ``Batch normalization: Accelerating deep network training by reducing
  internal covariate shift,''
\newblock in {\em International Conference on Machine Learning}, 2015, pp.
  448--456.

\bibitem{xu2015empirical}
Bing Xu, Naiyan Wang, Tianqi Chen, and Mu~Li,
\newblock ``Empirical evaluation of rectified activations in convolutional
  network,''
\newblock {\em arXiv preprint arXiv:1505.00853}, 2015.

\bibitem{kingmaadam}
Diederik~P Kingma and Jimmy~Lei Ba,
\newblock ``Adam: Amethod for stochastic optimization,''
\newblock .

\bibitem{panayotov2015librispeech}
Vassil Panayotov, Guoguo Chen, Daniel Povey, and Sanjeev Khudanpur,
\newblock ``Librispeech: an asr corpus based on public domain audio books,''
\newblock in {\em 2015 IEEE International Conference on Acoustics, Speech and
  Signal Processing (ICASSP)}. IEEE, 2015, pp. 5206--5210.

\bibitem{maaten2008visualizing}
Laurens van~der Maaten and Geoffrey Hinton,
\newblock ``Visualizing data using t-sne,''
\newblock {\em Journal of machine learning research}, vol. 9, no. Nov, pp.
  2579--2605, 2008.

\bibitem{wu2018improving}
Chunyang Wu, Mark~JF Gales, Anton Ragni, Penny Karanasou, and Khe~Chai Sim,
\newblock ``Improving interpretability and regularization in deep learning,''
\newblock {\em IEEE/ACM Transactions on Audio, Speech, and Language
  Processing}, vol. 26, no. 2, pp. 256--265, 2018.

\bibitem{list2010sca}
Johann-Mattis List,
\newblock ``Sca: phonetic alignment based on sound classes,''
\newblock in {\em New Directions in Logic, Language and Computation}, pp.
  32--51. Springer, 2010.

\end{thebibliography}

\end{document}